\documentclass{article}

\PassOptionsToPackage{hyphens}{url}
\PassOptionsToPackage{numbers}{natbib}
\PassOptionsToPackage{numbers, sort}{natbib}
\usepackage[dblblindworkshop, final]{neurips_2026}
\usepackage{amsmath}
\usepackage{booktabs}
\usepackage{graphicx}
\usepackage{pgfplots}
\pgfplotsset{compat=1.18} \usepgfplotslibrary{groupplots} \usetikzlibrary{arrows.meta, decorations.pathreplacing}
\usepackage{xcolor}
\usepackage{placeins}

\definecolor{cHelp}{HTML}{1F4E79}
\definecolor{cHurt}{HTML}{A02020}

\workshoptitle{The 6th Workshop on Mathematical Reasoning and AI}

\usepackage[utf8]{inputenc} 
\usepackage[T1]{fontenc}    
\usepackage{hyperref}       
\usepackage{url}            
\usepackage{booktabs}       
\usepackage{amsfonts}       
\usepackage{nicefrac}       
\usepackage{microtype}      
\usepackage{xcolor}         

\title{Attention-Graph Curvature for the \\Verification of Mathematical Reasoning}
\hypersetup{
  pdftitle={Attention-Graph Curvature for the Verification of Mathematical Reasoning},
  pdfauthor={Kanghyeon Kim, Johannes Lutzeyer, Yanzhu Guo}
}

\author{%
  Kanghyeon Kim \\
  T\'el\'ecom Paris, IPP \\
  \texttt{kanghyeon.kim@telecom-paris.fr}
  \And
  Johannes F. Lutzeyer \\
  LIX, CNRS, École Polytechnique, IPP \\
  \texttt{johannes.lutzeyer@polytechnique.edu}
  \And
  Yanzhu Guo \\
  T\'el\'ecom Paris, IPP \\
  \texttt{yanzhu.guo@telecom-paris.fr}
}

\begin{document}

\maketitle

\begin{abstract}
    Language models have recently solved open Erdős problems and achieved gold-medal level performance on International Mathematical Olympiad problems. But their solutions, written in natural language, cannot yet be verified by proof assistants. An erroneous reasoning step can invalidate every step that comes after, so it is advantageous to detect errors early. Several methods have been proposed to detect erroneous reasoning quickly based on the features of the forward pass of the language model that performs reasoning. In this work, we study whether the directed Forman curvature of that model's attention graph, a local quantity computed at an edge from the attention weights around it, carries information about a step's correctness. To do so, we evaluate four per-step verifiers on ProcessBench with and without the curvature appended to their input. Appending curvature values improves the overall mean performance of all four verifiers. ReProbe, the verifier that takes every token of a step as input, shows the largest improvement. Adding the curvature increases the time of the language model's forward pass by 18\% on average. Our code is available at \url{https://github.com/kanghyeonz/attention-graph-curvature}.
\end{abstract}

\section{Introduction}
\label{introduction}

Language models now solve mathematics problems of Olympiad and research difficulty. Gemini Deep Think scored 35 of 42 points at the 2025 International Mathematical Olympiad with solutions written in natural language \citep{deepmind2025imo}, and an agent built on Gemini Deep Think solved or partially solved four problems listed as open in the Erdős problems database \citep{feng2026semiautonomousmathematicsdiscoverygemini}. Language models, which we call \emph{generators}, write such solutions in natural language that no proof assistant can check, so an erroneous step has to be detected quickly and automatically. Methods that detect erroneous mathematical reasoning from features of the generator's forward pass have been proposed \citep{zhang2025reasoning,ni-etal-2026-reprobe,noel2026geometry}, and they use one of two types of features: 1) the hidden states, on which \citet{zhang2025reasoning} train an MLP probe to classify a  reasoning segment ending in an intermediate answer and \citet{ni-etal-2026-reprobe} train a transformer encoder to classify a reasoning step, and 2) the attention matrix taken as a weighted graph, the \emph{attention graph}, from which \citet{noel2026geometry} computes spectral statistics of the hidden states to classify a whole solution. We adapt all three methods to construct \textit{per-step verifiers} that score each step of a solution (Section~\ref{sec:setup}).

In this paper, we study whether the local geometry of the attention graph carries information about the correctness of a step, that is, whether a per-step verifier detects an erroneous step more accurately when given the local geometry. The directed Forman curvature \citep{sreejith2017formancurvaturedirectednetworks}, which generalizes the Forman curvature \citep{Forman2003BochnersMF} to directed graphs, is one quantity that describes this geometry at each edge. The curvature of an edge is a closed-form function of the attention weights of the edges that meet its two endpoints, so it requires no training and can be computed at every layer for every step (Section~\ref{sec:curv}). In particular, the curvature can reflect an attention connection that is weak relative to the connections entering its source and leaving its target, potentially acting as a bottleneck. Our intuition is that the distribution of these relative connection strengths across a step may help distinguish correct from erroneous reasoning. We therefore test whether appending curvature features to a verifier's input improves its ability to detect errors. Our results show that the directed Forman curvature provides information beyond what is already captured by the existing inputs of these per-step verifiers (Section~\ref{sec:pairs}).

\section{Methodology}
\label{sec:method}

We begin with the notation for a solution and its steps. Section~\ref{sec:curv} defines the curvature of the attention graph and the curvature vector $\mathbf{q}$ that we append to a verifier's input. Section~\ref{sec:setup} describes the four per-step verifiers.

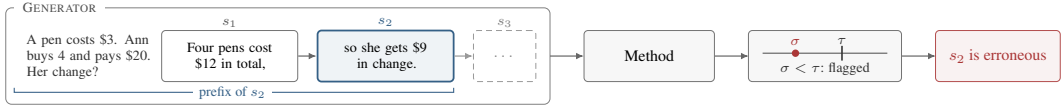
\begin{figure}[!t]
    \centering
    \scalebox{0.75}{\begin{tikzpicture}[
        font=\footnotesize, x=1cm, y=1cm,
        >={Latex[length=4pt,width=3.4pt]},
        box/.style={draw=black!55, rounded corners=2pt, line width=0.5pt, align=center,
        inner sep=3pt, minimum height=0.9cm, fill=white},
        sbox/.style={box, minimum width=2.4cm, text width=2.15cm, font=\scriptsize},
        cur/.style={sbox, draw=cHelp!90, line width=0.9pt, fill=cHelp!7},
        ghost/.style={sbox, minimum width=1.1cm, text width=0.8cm, draw=black!40, densely dashed, fill=none, text=black!50},
        unit/.style={box, minimum width=2.3cm, fill=black!3},
        idx/.style={font=\scriptsize, text=black!65, anchor=south, inner sep=1.5pt},
        edge/.style={->, draw=black!60, line width=0.6pt}]
        \draw[draw=black!50, rounded corners=3pt, line width=0.5pt] (0,-0.85) rectangle (9.6,0.85);
        \node[fill=white, inner xsep=4pt, inner ysep=1pt, text=black!70, font=\scriptsize] at (0.95,0.85) {\textsc{Generator}};
        \draw[draw=cHelp!80, line width=0.6pt] (0.15,-0.56) -- (0.15,-0.64) -- (7.9,-0.64) -- (7.9,-0.56);
        \node[fill=white, inner xsep=3pt, inner ysep=0.5pt, text=cHelp!90, font=\scriptsize] at (4.0,-0.64) {prefix of $s_{2}$};
        \node[anchor=west, text=black!80, font=\scriptsize, align=left, text width=2.3cm]
        at (0.25,0) {A pen costs \$3. Ann buys 4 and pays \$20. Her change?};
        \node[sbox]  (s1) at (3.95,0) {Four pens cost \$12 in total,};
        \node[cur]   (s2) at (6.70,0) {so she gets \$9 in change.};
        \node[ghost] (s3) at (8.80,0) {$\dots$};
        \draw[edge] (s1) -- (s2);
        \draw[edge, draw=black!40] (s2) -- (s3);
        \node[idx]                at (s1.north) {$s_1$};
        \node[idx, text=cHelp!90] at (s2.north) {$s_2$};
        \node[idx, text=black!50] at (s3.north) {$s_3$};
        \draw[edge] (9.6,0) -- (10.2,0);
        \node[unit] (met) at (11.35,0) {Method};
        \draw[edge] (met.east) -- ++(0.6,0);
        \node[unit, minimum width=2.7cm] (thr) at (14.45,0) {};
        \begin{scope}[shift={(14.45,0.05)}]
        \draw[black!60, line width=0.6pt] (-1.1,0) -- (1.1,0);
        \draw[black!80, line width=0.8pt] (0.3,-0.13) -- (0.3,0.13);
        \node[font=\scriptsize, text=black!80, anchor=south, inner sep=1.5pt] at (0.3,0.11) {$\tau$};
        \fill[cHurt!90] (-0.52,0) circle (0.065);
        \node[font=\scriptsize, text=cHurt!90, anchor=south, inner sep=1.5pt] at (-0.52,0.09) {$\sigma$};
        \node[font=\scriptsize, text=black!80, anchor=north, inner sep=2pt] at (0,-0.09) {$\sigma<\tau$: flagged};
        \end{scope}
        \draw[edge] (thr.east) -- ++(0.6,0);
        \node[box, minimum width=2.2cm, draw=cHurt!80, fill=cHurt!6, text=cHurt!90] at (17.5,0) {$s_2$ is erroneous};
    \end{tikzpicture}}
    \caption{Verifier framework and evaluation. A verifier maps features extracted from the generator's forward pass on a step's prefix to a score, and the rule in Section~\ref{sec:task} determines whether a step is flagged. An erroneous solution is correctly classified when the first flagged step is the annotated first erroneous step, and a correct solution when no step is flagged.}
    \label{fig:method}
\end{figure}

A solution consists of a sequence of steps $s_{1},\dots,s_{N}$, each corresponding to a paragraph of the solution. The problem $x$ followed by the steps of its solution gives the token sequence

\begin{equation}
    x,\;\underbrace{t_{s_{1},1},\dots,t_{s_{1},|s_{1}|}}_{s_{1}},\;\dots,\;\underbrace{t_{s_{N},1},\dots,t_{s_{N},|s_{N}|}}_{s_{N}},
    \label{eq:notation}
\end{equation}

in which $|s_{i}|$ is the number of tokens in step $s_{i}$ and $t_{s_{i},k}$ the $k$-th token of step $s_{i}$. The generator is frozen, and $h_{s_{i},k}\in\mathbb{R}^{m}$ denotes its last-layer hidden state at $t_{s_{i},k}$, with $m$ the hidden dimension and $L$ the layer count. Every solution is annotated in one of two ways (Section~\ref{sec:task}). An erroneous solution comes with the index of its first erroneous step, and we \emph{label} all preceding steps as correct and the indexed step as erroneous, leaving subsequent steps unlabeled. A correct solution has no erroneous-step index, and we label all of its steps as correct. Figure~\ref{fig:method} illustrates the verifier framework using an example solution.

\subsection{Curvature of the attention graph}
\label{sec:curv}

Let $T_{i}=|x|+\sum_{j=1}^{i}|s_{j}|$ denote the number of tokens from the first token of the problem up to and including the final token of step $s_{i}$, where $|x|$ is the number of tokens in the problem. At every layer $\ell$ we average the attention weights over heads and remove self-loops. This yields, for each step $s_{i}$, one weighted directed graph per layer on the first $T_{i}$ tokens, i.e., the prefix ending at $s_{i}$. An edge $e=(v_{1},v_{2})$ is included wherever query $v_{1}$ attends to key $v_{2}$ with a nonzero weight $a_{e}$ (Appendix~\ref{app:local}), and is assigned weight $w_{e}=\max(-\log a_{e},\varepsilon)$, with $\varepsilon=10^{-9}$. The directed Forman curvature \citep{sreejith2017formancurvaturedirectednetworks} of an edge, with unit vertex weights, is
\begin{equation}
    \mathbf{F}(v_{1},v_{2})=2-\sqrt{w_{e}}\left(\sum_{e'\in\mathcal{E}_{\text{in}}(v_{1})}\frac{1}{\sqrt{w_{e'}}}+\sum_{e'\in\mathcal{E}_{\text{out}}(v_{2})}\frac{1}{\sqrt{w_{e'}}}\right),
\label{eq:forman}
\end{equation}
where $\mathcal{E}_{\text{in}}(v)$ is the set of edges into $v$ and $\mathcal{E}_{\text{out}}(v)$ the set of edges out of $v$, both in the graph from the same layer as $e$. This curvature compares the attention on an edge with the attention on the edges entering its source and leaving its target. With the other weights fixed, weaker attention on the central edge or stronger attention on these neighboring edges cannot increase its curvature. It can therefore describe a connection that is weak relative to the connections on either side, which motivates our intuition of a local bottleneck. We hypothesize that the distribution of such local relations differs between correct and erroneous steps. The sign also depends on the number of neighboring edges, so negative curvature alone does not establish a bottleneck or a reasoning error (Appendix~\ref{app:local}).

\begin{table}[!t]
    \caption{Inputs to each method without and with the curvature vector, with $\boldsymbol{\phi}_{i}\in\mathbb{R}^{5L}$ the spectral statistics of Appendix~\ref{app:baselines} and $h_{s_{i},k}\in\mathbb{R}^{m}$ the last-layer hidden state.}
    \label{tab:curvin}
    \centering
    \footnotesize
    \begin{tabular}{@{}lll@{}}
        \toprule
        Method & Input & With curvature \\
        \midrule
        Spectral, logistic & $\boldsymbol{\phi}_{i}$, one vector per step & $\left(\boldsymbol{\phi}_{i},\;\mathbf{q}_{s_{i}}\right)\in\mathbb{R}^{10L}$ \\
        Hidden states, linear & $h_{s_{i},|s_{i}|}$, the final token & $\left(h_{s_{i},|s_{i}|},\;\mathbf{q}_{s_{i},|s_{i}|}\right)\in\mathbb{R}^{m+5L}$ \\
        MLP probe & $h_{s_{i},|s_{i}|}$, the final token & $\left(h_{s_{i},|s_{i}|},\;\mathbf{q}_{s_{i},|s_{i}|}\right)\in\mathbb{R}^{m+5L}$ \\
        ReProbe & $\left[h_{s_{i},k}\right]_{k=1}^{|s_{i}|}$, every token of the step & $\left[\left(h_{s_{i},k},\;\mathbf{q}_{s_{i},k}\right)\right]_{k=1}^{|s_{i}|}\in\mathbb{R}^{|s_{i}|\times(m+5L)}$ \\
        \bottomrule
    \end{tabular}
\end{table}

Curvature is defined on edges, whereas a per-step verifier takes tokens or steps as input, so we aggregate them accordingly. For a set $S$ of tokens, we consider the graph of the prefix that ends at the step containing them. At each layer $\ell$, we summarize the curvatures of their outgoing edges by their quantiles, and stack these quantiles across layers to obtain the curvature vector of $S$. Let $\mathcal{E}^{(\ell)}_{\text{out}}(S)=\bigcup_{t\in S}\mathcal{E}_{\text{out}}(t)$ denote the set of edges leaving tokens in $S$ at layer $\ell$ of the graph. Then
\begin{equation}
\begin{aligned}
    \mathbf{q}^{(\ell)}_{S}&=\left(Q_{\alpha}\left\{\mathbf{F}(e):e\in\mathcal{E}^{(\ell)}_{\text{out}}(S)\right\}\right)_{\alpha\in\{0,\,0.25,\,0.5,\,0.75,\,1\}}\in\mathbb{R}^{5},\\
    \mathbf{q}_{S}&=\left(\mathbf{q}^{(\ell)}_{S}\right)_{\ell=1}^{L}\in\mathbb{R}^{5L}.
\end{aligned}
\label{eq:q}
\end{equation}
Here $Q_{\alpha}$ denotes the $\alpha$-quantile of the curvatures of the edges in $\mathcal{E}^{(\ell)}_{\text{out}}(S)$. Two choices of $S$ are used. For a single token $t$ we write $\mathbf{q}_{t}$, and $\mathbf{q}_{s_{i},k}$ denotes $\mathbf{q}_{t}$ at the token $t_{s_{i},k}$, with the same subscripts as $h_{s_{i},k}$. For all the tokens of step $s_{i}$ we write $\mathbf{q}_{s_{i}}$. A method that takes hidden states receives $\mathbf{q}_{t}$ for each token whose hidden state it takes, whereas a method that takes one vector of statistics per step receives $\mathbf{q}_{s_{i}}$ (Table~\ref{tab:curvin}). Because causal attention only allows a token to attend to itself and preceding tokens, the prefix graph of step $s_{i}$ is fixed once $s_{i}$ has been generated. Therefore $\mathbf{q}$ can be computed and the step scored immediately. For a finished solution, one single forward pass provides the graphs for all steps. Appendix~\ref{app:local} gives the conditions under which the maximum-quantile coordinates are constant across steps in the final-token and pooled step representations.

\subsection{Per-step verifiers}
\label{sec:setup}

We compare four per-step verifiers, which we refer to as \textit{methods}, each specified in Appendix~\ref{app:baselines}, and Table~\ref{tab:curvin} states their inputs. \emph{Spectral, logistic} uses the five spectral statistics of \citet{noel2026geometry} at every layer, giving $5L$ features as input to a logistic classifier. While \citet{noel2026geometry} thresholds one statistic at one layer, we keep the statistics but replace the classifier so that this method and \emph{Hidden states, linear} differ only in their input. \emph{Hidden states, linear} applies the same logistic classifier to the last-layer hidden state at the final token of a step. The \emph{MLP probe} of \citet{zhang2025reasoning} is a two-layer network on the last-layer hidden state. We train it at the final token of a step to predict the step's correctness with the hyperparameter grid and selection rule of that paper. \emph{ReProbe} \citep{ni-etal-2026-reprobe} is a transformer encoder over the last-layer hidden states of the tokens within a step, trained with the hyperparameters that its paper reports.

To test whether the gains from curvature can also be obtained with simpler attention features, we append two alternative representations to each method's original input. The first takes the same five quantiles of the edge weights $w_e$ instead of their curvatures, giving $5L$ features (Neg-log). The second takes normalized attention entropy, concentration, and first-token attention mass at every layer, giving $3L$ features (Stats). Each representation is supplied at the same token or step resolution as curvature (Appendix~\ref{app:baselines}).
\newpage 
\section{Results}
\label{sec:results}

In this section we present our empirical results. Section~\ref{sec:task} describes the benchmark on which the methods are evaluated, the metric, and the paired comparison used to evaluate each method with and without the curvature vector. Section~\ref{sec:pairs} reports the paired differences and the comparisons with the two simpler attention representations, and Section~\ref{sec:cost} the cost of obtaining and appending the curvature vector.

\subsection{Benchmark and metric}
\label{sec:task}
\label{sec:design}

ProcessBench \citep{zheng-etal-2025-processbench} is a collection of step-by-step solutions written in natural language by open-weight language models for problems from four subsets: GSM8K, MATH, OlympiadBench and Omni-MATH. For every solution, human experts annotate either the first erroneous step or that no step is erroneous. We use the solutions from the seven generators with 8B parameters or fewer. ProcessBench has no OlympiadBench or Omni-MATH solutions from three of the generators, so the seven generators give a total of 22 \emph{generator--subset combinations}, each consisting of the solutions one generator produced for one subset. For each generator--subset combination, every method is trained and evaluated by stratified five-fold cross-validation grouped by problem, repeated over $10$ seeds. For test fold $i$, validation uses fold $(i+1)\bmod 5$ and fitting uses the other three folds, without refitting on validation data (Appendix~\ref{app:evaluation}).

A method's classifier maps its input for a step to a score $\sigma$, the probability that the step is correct, and a step is \emph{flagged} when $\sigma<\tau$ for a threshold $\tau$. The metric $A$ is the one ProcessBench reports as its F1 score, the harmonic mean of the fraction of correct solutions in which no step is flagged and the fraction of erroneous solutions whose first flagged step is the annotated first erroneous step. The $A$ of a generator--subset combination is the mean over the seeds of the value computed from that seed's pooled held-out decisions, and the threshold $\tau$ is the only quantity chosen to maximize $A$, on the selected model's own scores for its three fitting folds (Appendix~\ref{app:evaluation}). For each of the four methods, the configurations without and with curvature form a \emph{pair} with the same folds, seeds, classifier type, and threshold selection rule. The Neg-log and Stats configurations use those same folds and seeds, and the same normalization, fitting, and model-selection rules as the corresponding method. Each configuration selects its own threshold from its own fitting scores. The paired tests reported in Table~\ref{tab:pairs} compare curvature with the original input. We report the paired difference $\Delta A_{c}=A^{\text{with cur}}_{c}-A^{\text{w/o cur}}_{c}$ for each generator--subset combination $c$. We summarize the 22 differences using their solution-count-weighted mean, the number of positive differences, and a two-sided Wilcoxon signed-rank test (Appendix~\ref{app:evaluation}).

\subsection{Effects of curvature features}
\label{sec:pairs}

Table~\ref{tab:pairs} summarizes the four curvature pairs and the two attention controls for each method over the 22 generator--subset combinations. Appendix~\ref{app:detailed} reports $A$ and $\Delta A$ for every generator--subset combination and the mean of $A$ over the generator--subset combinations of each subset. 

Appending curvature features yields the highest overall mean performance for all four methods (Table~\ref{tab:pairs}). The simpler attention representations, Neg-log and Stats, also improve mean performance in most cases, but their gains are smaller than those obtained with curvature. The magnitude and consistency of the curvature gains nevertheless vary across verifiers.

ReProbe shows the largest improvement, with $\Delta A=+0.054$ and positive differences in 18 of the 22 generator--subset combinations with $p=0.0011$. Its performance also improves on every subset, with gains ranging from $+0.03$ on GSM8K to $+0.07$ on MATH (Table~\ref{tab:main}). The spectral logistic and hidden-state linear classifiers show smaller mean improvements of $+0.021$ and $+0.008$, with positive differences in 18 and 14 combinations, respectively; both improvements have $p<0.05$. The MLP probe shows the smallest mean improvement, at $+0.005$, with positive differences in 13 combinations, and its improvement is not statistically significant ($p=0.483$).

ReProbe's larger gains may be related to how it incorporates curvature. It receives $\mathbf{q}_{s_{i},k}$ at every token of a step (Table~\ref{tab:curvin}), so its input retains the information of how the curvature varies across the tokens of the step. In contrast, the three other methods receive a single vector per step, either from the final token or from quantiles computed over the step, which does not preserve the full token-level variation. We hypothesize that this variation carries useful information about a step's correctness, but leave a direct test of this hypothesis to future work.

\begin{table}[t]
    \caption{Each method with its original input (Base) or with one attention-derived representation appended: five quantiles of $w_e$ per layer (Neg-log, $5L$ features), normalized attention entropy, concentration, and first-token attention mass (Stats, $3L$), or curvature ($\mathbf{q}$, $5L$). All configurations use the same folds and seeds. Within each method, the fitting and threshold-selection rules are identical across feature configurations. $A$ is the harmonic mean of the accuracies on correct and erroneous solutions, averaged over the 22 generator--subset combinations with solution-count weights. $\Delta A$ and the unadjusted two-sided Wilcoxon $p$-value compare curvature with Base only (Section~\ref{sec:design}); they do not test curvature against Neg-log or Stats.}
    \label{tab:pairs}
    \centering
    \small
    \setlength{\tabcolsep}{4pt}
    \begin{tabular}{@{}lccccrr@{}}
        \toprule
        & \multicolumn{4}{c}{$A$} & \multicolumn{2}{c}{Curvature vs Base} \\
        \cmidrule(lr){2-5}\cmidrule(lr){6-7}
        Method & Base & + Neg-log & + Stats & + Curvature (ours) & $\Delta A$ & $p$ \\
        \midrule
        Spectral, logistic & 0.3261 & 0.3165 & 0.3298 & \textbf{0.3474} & $+0.0213$ & $<$0.001 \\
        Hidden states, linear & 0.1828 & 0.1872 & 0.1847 & \textbf{0.1907} & $+0.0079$ & 0.044 \\
        MLP probe & 0.2991 & 0.3033 & 0.3021 & \textbf{0.3037} & $+0.0047$ & 0.483 \\
        ReProbe & 0.2932 & 0.3068 & 0.3103 & \textbf{0.3472} & $+0.0540$ & 0.0011 \\
        \bottomrule
    \end{tabular}
\end{table}

\subsection{Computational cost}
\label{sec:cost}

Adding curvature introduces only modest computational overhead relative to the generator's forward pass. Returning the attention and averaging over heads add 14.6 ms to a 118 ms forward pass, while computing the curvature vector $\mathbf{q}$ takes 6.8 ms per step on average (Table~\ref{tab:cost}, Appendix~\ref{app:evaluation}). Once $\mathbf{q}$ is formed, appending it to a verifier increases the scoring time by less than 0.1 ms for all four methods. The main additional cost therefore comes from extracting and processing the attention graph rather than appending the curvature itself, and remains small compared with generation. Because the curvature vector can be computed as soon as a step is generated, this low overhead makes it practical to use curvature-based verification during inference to flag erroneous steps before generation continues.

\FloatBarrier
\section{Conclusion}
\label{sec:conclusion}

We introduced attention-graph curvature as a feature for verifying erroneous mathematical reasoning, to our knowledge the first use of directed Forman curvature for this purpose. Appending these curvature features increased the overall mean performance of all four methods, with the largest gain for ReProbe which receives token-level curvature. For ReProbe, curvature also gave larger gains than either of the two simpler attention-derived representations on all four subsets. We hypothesize that how curvature varies across tokens may carry useful information, and leave the direct comparison of token-level and step-level inputs to future work. The additional computational cost of curvature is relatively small compared with the generator's forward pass, and the features can be computed as soon as each step is generated. Future work can therefore use curvature-based verification to improve a generator's reasoning on the fly, for example by rejecting or resampling flagged steps or guiding generation toward alternative reasoning paths.

\begin{ack}
The work of Johannes F. Lutzeyer is supported by the French National Research Agency (ANR) via the ``GraspGNNs'' JCJC grant (ANR-24-CE23-3888).
\end{ack}
\newpage
\bibliographystyle{plainnat}
\bibliography{references}


\appendix

\section{Details of the curvature}
\label{app:local}

\paragraph{Choice of curvature.} The spectral analysis of attention graphs by \citet{noel2026geometry} motivates studying whether their local connectivity is also useful for verifying reasoning. The directed Forman curvature of an edge, Equation~\ref{eq:forman}, is a closed-form function of the attention weights of the edges that meet its two endpoints. It describes how the edge relates to its neighboring connections. The Ollivier--Ricci curvature \citep{Ollivier_2009} requires an optimal transport problem to be solved at every edge \citep{pmlr-v202-nguyen23c}, which is costly on a graph with one edge per attended token at every layer of every prefix, so we did not use it. This computational advantage motivates the choice but does not establish that Forman curvature is more predictive than other graph features.

\paragraph{Sign and magnitude.} The curvature of $e=(v_{1},v_{2})$ depends on the edges that arrive at its source $v_{1}$ and the edges that leave its target $v_{2}$. For fixed $w_e$, each of them lowers the curvature by an amount that is nondecreasing in its attention weight, since a larger attention weight gives a smaller $w$ until the floor $\varepsilon$ is reached. The factor $\sqrt{w_{e}}$ scales the sum, so with the neighboring weights fixed, decreasing the edge's attention cannot increase its curvature. The curvature can therefore reflect a connection that is weak relative to the incoming connections at its source and the outgoing connections at its target. Its sign also depends on the number of neighbors: if all weights in the expression are equal and the two sums contain a total of $k$ terms, the curvature is $2-k$. It can thus be negative even when the central edge is no weaker than its neighbors. Negative curvature alone does not identify a global bottleneck or a reasoning error. The connectivity interpretation in \citet{noel2026geometry} motivates this local analysis, but does not establish either implication for our directed curvature.

\paragraph{Edge weight.} Were the attention weights themselves taken as $w$, an attention weight of $10^{-6}$ would contribute a thousand times what a weight of $1$ does to the sums of Equation~\ref{eq:forman}, and softmax gives a query many entries of that size, so the sums would be dominated by the edges the model attends to least. We therefore take $w_{e}=\max(-\log a_{e},\varepsilon)$, and the floor $\varepsilon=10^{-9}$ prevents a division by zero in Equation~\ref{eq:forman} at an edge of attention weight exactly $1$, whose $-\log a_{e}$ is $0$. We retain all positive head-averaged attention entries except self-loops, without further sparsification or renormalization. The transformation is a modeling choice. The Neg-log control uses the same transformed weights without computing curvature, while Stats summarizes the attention weights directly (Appendix~\ref{app:baselines}).

\paragraph{Aggregation.} For each step, we restrict the graph to the prefix ending at that step before computing edge curvatures. Causal attention alone does not make this restriction unnecessary: later queries can add incoming edges to earlier tokens and change their curvature. A step's final token has no incoming edge within its prefix graph, so the summaries of Equation~\ref{eq:q} are taken over outgoing edges. A token with no outgoing edge, including the first token of the sequence after self-loops are removed, is assigned a zero vector. A quantile in Equation~\ref{eq:q} is computed by linear interpolation between adjacent order statistics. The step summary pools the outgoing edges of all its tokens before taking quantiles; it is not the average of the token summaries. These summaries describe distributions of curvature values, but do not preserve which edges connect which tokens. Using them across layers and, for ReProbe, across tokens lets the verifier use patterns involving multiple edges without treating an individual negative value as an error indicator.

\paragraph{Maximum quantile.} Every edge has curvature at most $2$. If a step's final token has a positive attention edge to the first token, that edge has curvature exactly $2$: its source has no incoming edge and its target has no outgoing edge, so both sums in Equation~\ref{eq:forman} vanish. Under this condition, the maximum quantile is $2$ at that layer both for the final token and for the pooled step edges. In finite precision, an attention entry may be zero, so the existence of this edge must be checked. The same argument does not apply to every token within the step, since earlier tokens can receive incoming edges from later tokens of that step. We retain all five quantiles. A coordinate that is constant on the fitting data becomes zero on those data after centering; removing the maximum quantile at every token would change ReProbe's representation beyond the removal of constant coordinates.

\section{Implementation details}
\label{app:baselines}

This appendix specifies the four methods in the notation of Equation~\ref{eq:notation}, and Table~\ref{tab:curvin} states what each input becomes with the curvature. The seven generators belong to the Qwen2, Qwen2.5, Qwen2.5-Math, Llama-3 and Llama-3.1 families. $L$ is 28 for the Qwen ones and 32 for the Llama ones, so $\mathbf{q}$ has 140 or 160 entries, and the hidden dimension $m$ is $1{,}536$ for the two 1.5B generators, $3{,}584$ for the three 7B Qwen generators, and $4{,}096$ for the two Llama generators. Each prior method was implemented from its paper and, where the authors released code, from that code as well, with the adaptations described below. All methods use the same stored generator features and fold assignments. Normalization statistics and class-frequency ratios are estimated from the three fitting folds; ReProbe's erroneous-class weight is fixed at $3$. The designated validation fold is used to select neural-network settings and checkpoints (Appendix~\ref{app:evaluation}). For every method the score $\sigma$ of a step is the probability of the correct-step class, and since ReProbe's head outputs the probability of the erroneous class, its $\sigma$ is one minus that output. The logistic classifier of \emph{Spectral, logistic} and \emph{Hidden states, linear} is per-feature standardization, with mean and standard deviation estimated on the training folds alone, followed by $\ell_{2}$-penalized logistic regression at the \texttt{scikit-learn} defaults ($C=1$, lbfgs, tolerance $10^{-4}$), with \texttt{max\_iter} increased to $2000$ and \texttt{class\_weight=\textquotesingle balanced\textquotesingle}. These settings are fixed across folds; validation is not used to tune the logistic classifiers.

\paragraph{Spectral, logistic.} The input is the five spectral statistics of \citet{noel2026geometry} at every layer, the Dirichlet energy, the spectral entropy, the high-frequency energy ratio, the Fiedler value, and a smoothness measure, $5L$ values, computed on the prefix graph with the heads uniformly averaged and the resulting matrix symmetrized, the combinatorial Laplacian, the median eigenvalue index as the high-frequency cutoff, and the smoothness in the form of the released library, the Dirichlet energy divided by the squared Frobenius norm of the layer's hidden states. The classifier of that paper thresholds one statistic at one layer, and we give all $5L$ values to the logistic classifier so that this method and \emph{Hidden states, linear} differ only in their input.

\paragraph{Hidden states, linear.} The input is $h_{s_{i},|s_{i}|}$, under the same logistic classifier.

\paragraph{MLP probe.} A two-layer perceptron with a ReLU hidden layer of width $d\in\{0,16,32\}$, with $d=0$ a single linear layer, on $h_{s_{i},|s_{i}|}$, following \citet{zhang2025reasoning}. We standardize each input dimension on the three fitting folds' labeled steps, using the sample standard deviation with a lower bound of $10^{-3}$. The objective is the cross-entropy with the correct class weighted by $\rho\gamma$, where $\rho$ is the ratio of erroneous to correct training steps, optimized with Adam at batch $64$ for at most $200$ epochs, with patience $10$ on the designated validation fold and the best epoch restored. An improvement must reduce validation loss by more than $10^{-6}$. Validation loss is averaged over labeled steps, weighting each batch by its number of steps. Four hyperparameters are searched on a grid of $216$ hyperparameter settings, the learning rate over $\{10^{-3},10^{-4},10^{-5}\}$, $\gamma$ over $\{0.3,0.5,0.7,0.9,1.0,1.5,2.0,3.0\}$, the weight decay over $\{10^{-3},10^{-2},10^{-1}\}$, and $d$. The 10 best by validation step accuracy at score threshold $0.5$ are retained and the smallest $d$ among them chosen, ties by validation accuracy. The selected setting is trained again with the same seed and fitting data, using the same validation checkpoint rule. Validation data are not added to the fitting set. \citet{zhang2025reasoning} train the probe at the end of a segment of the reasoning that ends in an intermediate answer, which they call a chunk, to predict whether that answer is correct, and we train it at a step's final token to predict the step's correctness. The patience, the tie-break, and the standardization of the inputs are ours, and the standardization interacts with weight decay and early stopping, so the selected hyperparameters are not comparable to that paper's.

\paragraph{ReProbe.} A two-layer projection to $\mathbb{R}^{512}$ of each token's $h_{s_{i},k}$, a learned marker embedding added to every token of the step, one encoder layer with 16 heads and dropout $0.1$, the mean over the step's tokens, and a two-layer head, following \citet{ni-etal-2026-reprobe}, with padded positions masked out of the encoder's attention. We standardize each input dimension on all retained tokens of the first 32 fitting solutions in the fixed data order, or all fitting solutions if fewer are available, with population standard deviations bounded below by $10^{-3}$. Training uses the learning rate of \citet{ni-etal-2026-reprobe}, $5\times10^{-4}$, their batch size of $128$, their weight of $3$ on the erroneous class, and their five epochs, with AdamW at weight decay $0.1$, linear warm-up over the first 5\% of the updates then linear decay to zero, and gradients clipped to norm $1$, and keeps the checkpoint with the lowest validation loss on the designated validation fold. The warm-up lasts at least one update. Validation loss is averaged over labeled steps, weighting batches by their numbers of steps. To limit memory use, each logical batch of up to 128 steps is processed in smaller batches, accumulating gradients with weights proportional to their sizes before one optimizer update. No further validation split is made. Our encoder receives only the current step's tokens; their generator hidden states already incorporate the preceding prefix. We use the last-layer hidden states, a maximum generator input length of $2048$ tokens, and the normalization and checkpoint rules specified here. These choices define our adaptation to the per-step task and the available data.

\paragraph{Attention controls.} Both controls use the same positive head-averaged attention entries and prefix graphs as curvature, with self-loops removed. They are computed at every layer. The retained input tokens and step boundaries are identical across configurations.

\paragraph{Negative-log attention.} For each token, we take the five quantiles of $w_e=\max(-\log a_e,10^{-9})$ over its outgoing edges, at the levels in Equation~\ref{eq:q}, using linear interpolation. The step-level representation pools the outgoing edge weights of all retained tokens in the step before taking quantiles. No attention renormalization is applied. An empty edge set gives five zeros. Stacking across layers gives $5L$ features, the same dimension as curvature.

\paragraph{Attention statistics.} At a fixed layer, let $K_t$ be the set of keys with positive attention from token $t$, excluding $t$ itself, and let $n_t=|K_t|$. For $n_t>0$, we normalize the retained outgoing weights as $p_{tv}=a_{tv}/\sum_{u\in K_t}a_{tu}$. For $n_t>1$, normalized entropy and concentration are
\begin{equation*}
    H_t=-\frac{\sum_{v\in K_t}p_{tv}\log p_{tv}}{\log n_t},
    \qquad
    C_t=\frac{\sum_{v\in K_t}p_{tv}^{2}-1/n_t}{1-1/n_t}.
\end{equation*}
The third statistic is $B_t=p_{t v_0}$, where $v_0$ is the first token of the full input, with $B_t=0$ if $v_0\notin K_t$. For $n_t=1$, we set $H_t=0$ and $C_t=1$; for $n_t=0$, all three statistics are zero. Numerical values are clipped to $[0,1]$. The step representation averages each of the three statistics over retained tokens in that step. Stacking across layers gives $3L$ features.

\paragraph{Control inputs and fitting.} The spectral classifier receives each control's step representation; the hidden-state linear classifier and MLP receive the control representation at the step's final retained token; ReProbe receives its representation at every retained token. In each case it is appended to the method's original input in place of curvature. The normalization, class weighting, hyperparameter grid where applicable, checkpoint selection, and threshold-selection rules are unchanged. Models are fitted separately for each configuration. The Neg-log comparison matches curvature's added dimension and input resolution. Stats has fewer features and is not a comparison at equal input dimension.

\paragraph{Numerical reproducibility.} Neural fitting uses fixed seeds and strict deterministic PyTorch algorithms, with cuDNN benchmarking and TF32 disabled. CUDA evaluation sets \texttt{CUBLAS\_WORKSPACE\_CONFIG=:4096:8} before execution. The features, partitions, software versions, and fitting outputs are recorded. Separate checks replay threshold selection and solution counts from saved scores, and compare repeated fits with unchanged and altered test labels. These checks concern the recorded execution environment, not equality across hardware or software versions.

\section{Cross-validation, threshold selection, and cost}
\label{app:evaluation}

The 22 generator--subset combinations contain $1{,}994$ solutions and $14{,}906$ steps, and a generator--subset combination contains 22 to 164 solutions and 118 to $1{,}436$ steps. ProcessBench's solutions come divided into steps at paragraph breaks that the benchmark inserted by reformatting, discarding the few solutions whose final answer the reformatting changed. Every input of a generator--subset combination comes from a forward pass of its own generator, its folds, thresholds, and metric are computed within it, and where generator--subset combinations are averaged the weight is each one's solution count. Each generator--subset combination is evaluated by stratified group cross-validation with five folds and seeds $0$ through $9$, a seed determining one partition of the solutions into folds. Solutions of the same problem remain in one fold. Stratification approximately balances correct and erroneous solutions, with problem groups taking precedence over exact class proportions and fold sizes. The partitions are fixed before evaluation and shared by all sixteen method--feature configurations, including the eight configurations using Neg-log or Stats.

For test fold $i\in\{0,\ldots,4\}$, validation uses fold $(i+1)\bmod 5$ and fitting uses the remaining three folds. The nominal proportions are $60\%$ fitting, $20\%$ validation, and $20\%$ test, with deviations due to integer counts and problem groups. The validation fold selects neural-network settings and checkpoints. Test data are excluded from fitting, normalization, checkpoint selection, and threshold selection. The selected model is not refitted on the combined fitting and validation folds.

For each test fold, the selected model scores its own three fitting folds, and $\tau$ is selected on these scores and applied to that model's test-fold scores. No scores from other fold models are used. The candidates are the quantiles of the step scores pooled over the fitting folds at the 50 levels $0,\tfrac{1}{49},\tfrac{2}{49},\dots,1$, taken after non-finite scores are removed and with duplicate quantile values removed. The classifiers are trained on labeled steps with cross-entropy objectives, so $A$ is not their training objective, and $\tau$ is the only quantity chosen to maximize it. The candidate maximizing $A$ on the fitting folds is kept, with ties broken in favor of the smallest $\tau$. Candidate values of $A$ are compared using integer success and total counts to avoid numerical tie-breaking differences. $A$ is zero when both class accuracies are zero and undefined when a class is absent. If the fitting set lacks either solution class or has no finite scores, the predefined fallback is $\tau=0.5$.

The threshold is selected on the same data used to fit the model, rather than on an independent calibration set, and can overfit those scores. The test folds evaluate the complete procedure, including this threshold-selection rule. Validation and test scores are not used to adjust the threshold candidates or their selection.

These out-of-fold decisions are pooled across the five folds before the metric is computed. Let $\mathcal{S}_{\text{correct}}$ and $\mathcal{S}_{\text{err}}$ denote the correct and erroneous solutions of the generator--subset combination, and $\hat{\kappa}^{(r)}_{n}$ the index of the first step flagged in solution $n$ under seed $r$ ($\emptyset$ if none),
\begin{align}
    A^{(r)}_{\text{correct}}=\frac{1}{|\mathcal{S}_{\text{correct}}|}\sum_{n\in\mathcal{S}_{\text{correct}}}\mathbf{1}\!\left[\hat{\kappa}^{(r)}_{n}=\emptyset\right],
    \qquad
    A^{(r)}_{\text{err}}=\frac{1}{|\mathcal{S}_{\text{err}}|}\sum_{n\in\mathcal{S}_{\text{err}}}\mathbf{1}\!\left[\hat{\kappa}^{(r)}_{n}=\kappa^{\star}_{n}\right],
\end{align}
with $\kappa^{\star}_{n}$ the annotated first erroneous step, and $A^{(r)}$ their harmonic mean, taken as $0$ when both ratios are $0$. Here $1-A^{(r)}_{\text{correct}}$ is the false-positive rate among fully correct solutions, not among individual correct steps, and $A^{(r)}_{\text{err}}$ measures exact first-error localization. The reported value and standard deviation are then
\begin{align}
    \mu=\frac{1}{10}\sum_{r=0}^{9}A^{(r)},
    \qquad
    \mathrm{sd}=\sqrt{\frac{1}{9}\sum_{r=0}^{9}\left(A^{(r)}-\mu\right)^{2}}.
\end{align}
$A$ is formed once per seed from the pooled decisions rather than averaged over the five per-fold values: averaging harmonic means gives a different statistic, and a fold containing no solution of one class has no defined value, by analogy with the F-measure \citep{forman2010apples}. The standard deviation is taken over the 10 seeds and not over the 50 folds, since folds within one seed share training data \citep{NIPS2003_e82c4b19}. It describes variation across partitions and training seeds on the same data, not a standard error from independent datasets. The signed-rank test of Section~\ref{sec:design} uses the 22 combination-level mean differences between curvature and the original input, leaving out exact zeros. We report unadjusted two-sided p-values. Comparisons with Neg-log and Stats are descriptive; these tests do not assess their differences from curvature. Combinations can share benchmark problems across generators, so they should not be interpreted as 22 independent samples of problems.

\paragraph{Small samples.} The GSM8K/Qwen2-1.5B combination has 41 solutions, only four of them fully correct. Five folds therefore cannot each contain a fully correct solution. A validation or test fold may lack this class even when its labeled steps contain both classes, since an erroneous solution can contain correct steps before its first error. Step-level validation loss remains defined, but cannot assess acceptance of fully correct solutions in that fold. We retain the fixed fold rotation and pool test counts before computing $A$. Small validation sets can still make checkpoint selection unstable.

\paragraph{Truncation.} Solutions are tokenized with truncation at $2048$ tokens. A partially retained step is scored from its retained tokens, using the final retained token for the two final-token verifiers. A step with no retained tokens is excluded from fitting and validation losses and is treated as not flagged by every method, including the spectral classifier. Its solution remains in the metric denominator, so a first error beyond the retained prefix cannot be detected under this rule. There are eight steps with no retained tokens in four solutions of the 22 generator--subset combinations.

\paragraph{Timings.} All timings are on one NVIDIA GeForce RTX 3090 with 24 GB of memory and an AMD EPYC 7232P CPU, on random matrices and random token sequences at the prefix lengths of the scored steps, and Table~\ref{tab:cost} averages them over the steps of each generator--subset combination, from timings on a grid of prefix lengths interpolated at each step's own, and over the generator--subset combinations of a subset weighted by solution count. The generator's plain forward pass uses a fused kernel and returns no attention matrices, whereas a method that takes the attention graph as input requires a pass that returns them. At the median prefix that pass takes 2 to 16 ms more than the plain pass, whose own time is 27 to 150 ms, and the mean over heads takes under 1 ms. At the longest prefix the difference is 6 to 63 ms where the pass runs within memory, and in 12 of the 22 generator--subset combinations, those of the 7B and 8B generators with prefixes above about $1{,}500$ tokens, the matrices of every layer and head together with the model exceed the 24 GB, so Table~\ref{tab:cost} averages over the steps within memory, 98 percent of them. These measurements use synthetic inputs and exclude prefixes that exceed the memory limit; they do not measure the latency of end-to-end generation with verification. Computing $\mathbf{q}$ for the longest prefix of a generator--subset combination takes 3 to 72 ms at a peak of 266 to $7{,}937$ MiB, against 2 to 6 ms and 61 to 577 MiB at the median prefix, computed for all layers at once. Scoring one step peaks at 9.6 MiB of GPU memory for the MLP probe and 32 to 109 MiB for ReProbe, which takes at most 1.1 ms at a generator--subset combination's longest step, and the logistic classifiers use no GPU. With $\mathbf{q}$ appended, each method's scoring time changes by less than 0.1 ms, at most 0.07 ms for ReProbe.

\begin{table}[htbp]
    \caption{Cost per step of what the curvature vector requires, in ms, averaged over the prefix lengths of the scored steps of each generator--subset combination and then over the generator--subset combinations of a subset weighted by solution count. \emph{Plain} is the generator's forward pass with fused attention, which every method requires; \emph{attention} is what the pass that returns the attention matrices adds to it, and \emph{head mean} the mean over heads at every layer; \emph{forming $\mathbf{q}$} is the edge weight, the curvature of Equation~\ref{eq:forman}, and the quantiles of Equation~\ref{eq:q} at every layer, computed for all layers at once. \emph{Within memory} is the share of steps whose prefix the pass returning the attention matrices completes within the 24 GB of the GPU; the other entries average over those steps, and \emph{forming $\mathbf{q}$} over all steps. Timings on random matrices and random tokens at the steps' prefix lengths (Appendix~\ref{app:evaluation}).}
    \label{tab:cost}
    \centering
    \small
    \begin{tabular}{@{}lrrrrr@{}}
        \toprule
        Subset & Plain & Attention & Head mean & Forming $\mathbf{q}$ & Within memory \\
        \midrule
        GSM8K & 60 & 5.5 & 0.5 & 2.4 & 100\% \\
        MATH & 102 & 11.6 & 0.7 & 5.7 & 99\% \\
        OlympiadBench & 145 & 17.8 & 1.0 & 8.7 & 97\% \\
        Omni-MATH & 139 & 16.5 & 1.0 & 8.3 & 97\% \\
        \midrule
        All & 118 & 13.7 & 0.9 & 6.8 & 98\% \\
        \bottomrule
    \end{tabular}
\end{table}

\section{Detailed results}
\label{app:detailed}

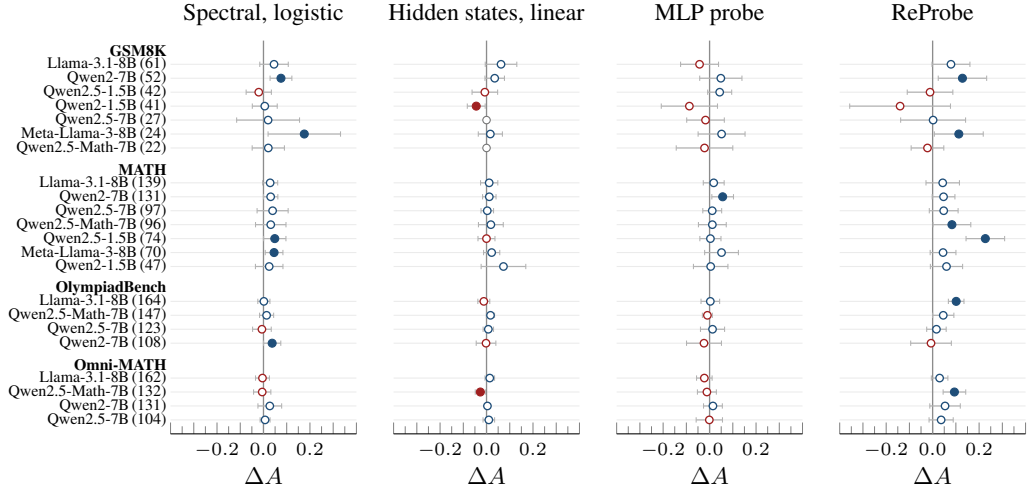
\begin{figure}[t]
  \centering
  \pgfplotsset{
    delta axis/.style={
      width=0.176\textwidth, height=5.2cm, scale only axis,
      xmin=-0.40, xmax=0.40, ymin=-0.75, ymax=27.45, y dir=reverse,
      axis x line*=bottom, axis y line*=left,
      y axis line style={draw=none}, ytick style={draw=none},
      xtick={-0.4,-0.3,-0.2,-0.1,0.0,0.1,0.2,0.3,0.4}, xticklabel style={font=\scriptsize},
      xticklabels={,,$-0.2$,,$0.0$,,$0.2$,,},
      minor xtick={-0.35,-0.25,-0.15,-0.05,0.05,0.15,0.25,0.35},
      xlabel={$\Delta A$}, xlabel style={font=\small, yshift=3pt},
      ytick={0.00,0.95,1.95,2.95,3.95,4.95,5.95,6.95,8.50,9.45,10.45,11.45,12.45,13.45,14.45,15.45,17.00,17.95,18.95,19.95,20.95,22.50,23.45,24.45,25.45,26.45},
      yticklabels={\textbf{GSM8K},Llama-3.1-8B\,(61),Qwen2-7B\,(52),Qwen2.5-1.5B\,(42),Qwen2-1.5B\,(41),Qwen2.5-7B\,(27),Meta-Llama-3-8B\,(24),Qwen2.5-Math-7B\,(22),\textbf{MATH},Llama-3.1-8B\,(139),Qwen2-7B\,(131),Qwen2.5-7B\,(97),Qwen2.5-Math-7B\,(96),Qwen2.5-1.5B\,(74),Meta-Llama-3-8B\,(70),Qwen2-1.5B\,(47),\textbf{OlympiadBench},Llama-3.1-8B\,(164),Qwen2.5-Math-7B\,(147),Qwen2.5-7B\,(123),Qwen2-7B\,(108),\textbf{Omni-MATH},Llama-3.1-8B\,(162),Qwen2.5-Math-7B\,(132),Qwen2-7B\,(131),Qwen2.5-7B\,(104)},
      yticklabel style={font=\tiny, xshift=2pt},
      title style={font=\small, yshift=-3pt, text height=1.5ex, text depth=.25ex}, clip=false,
    },
    dHelpF/.style={mark=*, mark size=1.5pt, draw=cHelp, mark options={fill=cHelp, draw=cHelp}},
    dHelpO/.style={mark=*, mark size=1.4pt, draw=cHelp, mark options={fill=white, draw=cHelp, line width=0.5pt}},
    dHurtF/.style={mark=*, mark size=1.5pt, draw=cHurt, mark options={fill=cHurt, draw=cHurt}},
    dHurtO/.style={mark=*, mark size=1.4pt, draw=cHurt, mark options={fill=white, draw=cHurt, line width=0.5pt}},
    dZero/.style={mark=*, mark size=1.4pt, draw=black!55, mark options={fill=white}},
    delta marks/.style={only marks, error bars/.cd, x dir=both, x explicit,
      error bar style={line width=0.45pt, draw=black!30},
      error mark options={rotate=90, mark size=1.0pt, draw=black!30}},
  }
  \begin{tikzpicture}
    \begin{groupplot}[group style={group size=4 by 1, horizontal sep=14pt, yticklabels at=edge left}, delta axis]
    \nextgroupplot[title={Spectral, logistic}]
      \draw[black!8, line width=0.4pt] (axis cs:-0.40,0.95) -- (axis cs:0.40,0.95);
      \draw[black!8, line width=0.4pt] (axis cs:-0.40,1.95) -- (axis cs:0.40,1.95);
      \draw[black!8, line width=0.4pt] (axis cs:-0.40,2.95) -- (axis cs:0.40,2.95);
      \draw[black!8, line width=0.4pt] (axis cs:-0.40,3.95) -- (axis cs:0.40,3.95);
      \draw[black!8, line width=0.4pt] (axis cs:-0.40,4.95) -- (axis cs:0.40,4.95);
      \draw[black!8, line width=0.4pt] (axis cs:-0.40,5.95) -- (axis cs:0.40,5.95);
      \draw[black!8, line width=0.4pt] (axis cs:-0.40,6.95) -- (axis cs:0.40,6.95);
      \draw[black!8, line width=0.4pt] (axis cs:-0.40,9.45) -- (axis cs:0.40,9.45);
      \draw[black!8, line width=0.4pt] (axis cs:-0.40,10.45) -- (axis cs:0.40,10.45);
      \draw[black!8, line width=0.4pt] (axis cs:-0.40,11.45) -- (axis cs:0.40,11.45);
      \draw[black!8, line width=0.4pt] (axis cs:-0.40,12.45) -- (axis cs:0.40,12.45);
      \draw[black!8, line width=0.4pt] (axis cs:-0.40,13.45) -- (axis cs:0.40,13.45);
      \draw[black!8, line width=0.4pt] (axis cs:-0.40,14.45) -- (axis cs:0.40,14.45);
      \draw[black!8, line width=0.4pt] (axis cs:-0.40,15.45) -- (axis cs:0.40,15.45);
      \draw[black!8, line width=0.4pt] (axis cs:-0.40,17.95) -- (axis cs:0.40,17.95);
      \draw[black!8, line width=0.4pt] (axis cs:-0.40,18.95) -- (axis cs:0.40,18.95);
      \draw[black!8, line width=0.4pt] (axis cs:-0.40,19.95) -- (axis cs:0.40,19.95);
      \draw[black!8, line width=0.4pt] (axis cs:-0.40,20.95) -- (axis cs:0.40,20.95);
      \draw[black!8, line width=0.4pt] (axis cs:-0.40,23.45) -- (axis cs:0.40,23.45);
      \draw[black!8, line width=0.4pt] (axis cs:-0.40,24.45) -- (axis cs:0.40,24.45);
      \draw[black!8, line width=0.4pt] (axis cs:-0.40,25.45) -- (axis cs:0.40,25.45);
      \draw[black!8, line width=0.4pt] (axis cs:-0.40,26.45) -- (axis cs:0.40,26.45);
      \draw[black!45, line width=0.5pt] (axis cs:0,-0.75) -- (axis cs:0,27.45);
      \addplot[dHelpF, delta marks] table[x index=0, y index=1, x error index=2] {
        0.07530 1.95 0.04688
        0.17568 5.95 0.15591
        0.04861 13.45 0.04808
        0.04564 14.45 0.03808
        0.03734 20.95 0.03717
      };
      \addplot[dHelpO, delta marks] table[x index=0, y index=1, x error index=2] {
        0.04541 0.95 0.06135
        0.00558 3.95 0.05365
        0.01983 4.95 0.13534
        0.02076 6.95 0.06940
        0.02870 9.45 0.03253
        0.03092 10.45 0.03094
        0.03933 11.45 0.06686
        0.03132 12.45 0.06558
        0.02460 15.45 0.05923
        0.00171 17.95 0.02609
        0.01346 18.95 0.03028
        0.02719 25.45 0.05104
        0.00691 26.45 0.02141
      };
      \addplot[dHurtO, delta marks] table[x index=0, y index=1, x error index=2] {
        -0.02033 2.95 0.05433
        -0.00669 19.95 0.04020
        -0.00434 23.45 0.02992
        -0.00544 24.45 0.03718
      };
    \nextgroupplot[title={Hidden states, linear}]
      \draw[black!8, line width=0.4pt] (axis cs:-0.40,0.95) -- (axis cs:0.40,0.95);
      \draw[black!8, line width=0.4pt] (axis cs:-0.40,1.95) -- (axis cs:0.40,1.95);
      \draw[black!8, line width=0.4pt] (axis cs:-0.40,2.95) -- (axis cs:0.40,2.95);
      \draw[black!8, line width=0.4pt] (axis cs:-0.40,3.95) -- (axis cs:0.40,3.95);
      \draw[black!8, line width=0.4pt] (axis cs:-0.40,4.95) -- (axis cs:0.40,4.95);
      \draw[black!8, line width=0.4pt] (axis cs:-0.40,5.95) -- (axis cs:0.40,5.95);
      \draw[black!8, line width=0.4pt] (axis cs:-0.40,6.95) -- (axis cs:0.40,6.95);
      \draw[black!8, line width=0.4pt] (axis cs:-0.40,9.45) -- (axis cs:0.40,9.45);
      \draw[black!8, line width=0.4pt] (axis cs:-0.40,10.45) -- (axis cs:0.40,10.45);
      \draw[black!8, line width=0.4pt] (axis cs:-0.40,11.45) -- (axis cs:0.40,11.45);
      \draw[black!8, line width=0.4pt] (axis cs:-0.40,12.45) -- (axis cs:0.40,12.45);
      \draw[black!8, line width=0.4pt] (axis cs:-0.40,13.45) -- (axis cs:0.40,13.45);
      \draw[black!8, line width=0.4pt] (axis cs:-0.40,14.45) -- (axis cs:0.40,14.45);
      \draw[black!8, line width=0.4pt] (axis cs:-0.40,15.45) -- (axis cs:0.40,15.45);
      \draw[black!8, line width=0.4pt] (axis cs:-0.40,17.95) -- (axis cs:0.40,17.95);
      \draw[black!8, line width=0.4pt] (axis cs:-0.40,18.95) -- (axis cs:0.40,18.95);
      \draw[black!8, line width=0.4pt] (axis cs:-0.40,19.95) -- (axis cs:0.40,19.95);
      \draw[black!8, line width=0.4pt] (axis cs:-0.40,20.95) -- (axis cs:0.40,20.95);
      \draw[black!8, line width=0.4pt] (axis cs:-0.40,23.45) -- (axis cs:0.40,23.45);
      \draw[black!8, line width=0.4pt] (axis cs:-0.40,24.45) -- (axis cs:0.40,24.45);
      \draw[black!8, line width=0.4pt] (axis cs:-0.40,25.45) -- (axis cs:0.40,25.45);
      \draw[black!8, line width=0.4pt] (axis cs:-0.40,26.45) -- (axis cs:0.40,26.45);
      \draw[black!45, line width=0.5pt] (axis cs:0,-0.75) -- (axis cs:0,27.45);
      \addplot[dHelpO, delta marks] table[x index=0, y index=1, x error index=2] {
        0.06239 0.95 0.06790
        0.03520 1.95 0.04196
        0.01642 5.95 0.05192
        0.01140 9.45 0.03720
        0.01193 10.45 0.02899
        0.00319 11.45 0.02701
        0.01863 12.45 0.05340
        0.02207 14.45 0.03489
        0.07287 15.45 0.09612
        0.01704 18.95 0.01754
        0.00766 19.95 0.02239
        0.01392 23.45 0.02061
        0.00386 25.45 0.01191
        0.00986 26.45 0.02487
      };
      \addplot[dHurtF, delta marks] table[x index=0, y index=1, x error index=2] {
        -0.04430 3.95 0.03849
        -0.02663 24.45 0.02188
      };
      \addplot[dHurtO, delta marks] table[x index=0, y index=1, x error index=2] {
        -0.00725 2.95 0.05504
        -0.00025 13.45 0.03655
        -0.01171 17.95 0.02567
        -0.00214 20.95 0.04230
      };
      \addplot[dZero, delta marks] table[x index=0, y index=1, x error index=2] {
        0.00000 4.95 0.00000
        0.00000 6.95 0.00000
      };
    \nextgroupplot[title={MLP probe}]
      \draw[black!8, line width=0.4pt] (axis cs:-0.40,0.95) -- (axis cs:0.40,0.95);
      \draw[black!8, line width=0.4pt] (axis cs:-0.40,1.95) -- (axis cs:0.40,1.95);
      \draw[black!8, line width=0.4pt] (axis cs:-0.40,2.95) -- (axis cs:0.40,2.95);
      \draw[black!8, line width=0.4pt] (axis cs:-0.40,3.95) -- (axis cs:0.40,3.95);
      \draw[black!8, line width=0.4pt] (axis cs:-0.40,4.95) -- (axis cs:0.40,4.95);
      \draw[black!8, line width=0.4pt] (axis cs:-0.40,5.95) -- (axis cs:0.40,5.95);
      \draw[black!8, line width=0.4pt] (axis cs:-0.40,6.95) -- (axis cs:0.40,6.95);
      \draw[black!8, line width=0.4pt] (axis cs:-0.40,9.45) -- (axis cs:0.40,9.45);
      \draw[black!8, line width=0.4pt] (axis cs:-0.40,10.45) -- (axis cs:0.40,10.45);
      \draw[black!8, line width=0.4pt] (axis cs:-0.40,11.45) -- (axis cs:0.40,11.45);
      \draw[black!8, line width=0.4pt] (axis cs:-0.40,12.45) -- (axis cs:0.40,12.45);
      \draw[black!8, line width=0.4pt] (axis cs:-0.40,13.45) -- (axis cs:0.40,13.45);
      \draw[black!8, line width=0.4pt] (axis cs:-0.40,14.45) -- (axis cs:0.40,14.45);
      \draw[black!8, line width=0.4pt] (axis cs:-0.40,15.45) -- (axis cs:0.40,15.45);
      \draw[black!8, line width=0.4pt] (axis cs:-0.40,17.95) -- (axis cs:0.40,17.95);
      \draw[black!8, line width=0.4pt] (axis cs:-0.40,18.95) -- (axis cs:0.40,18.95);
      \draw[black!8, line width=0.4pt] (axis cs:-0.40,19.95) -- (axis cs:0.40,19.95);
      \draw[black!8, line width=0.4pt] (axis cs:-0.40,20.95) -- (axis cs:0.40,20.95);
      \draw[black!8, line width=0.4pt] (axis cs:-0.40,23.45) -- (axis cs:0.40,23.45);
      \draw[black!8, line width=0.4pt] (axis cs:-0.40,24.45) -- (axis cs:0.40,24.45);
      \draw[black!8, line width=0.4pt] (axis cs:-0.40,25.45) -- (axis cs:0.40,25.45);
      \draw[black!8, line width=0.4pt] (axis cs:-0.40,26.45) -- (axis cs:0.40,26.45);
      \draw[black!45, line width=0.5pt] (axis cs:0,-0.75) -- (axis cs:0,27.45);
      \addplot[dHelpF, delta marks] table[x index=0, y index=1, x error index=2] {
        0.05627 10.45 0.04661
      };
      \addplot[dHelpO, delta marks] table[x index=0, y index=1, x error index=2] {
        0.04840 1.95 0.09093
        0.04351 2.95 0.05165
        0.05165 5.95 0.10104
        0.01779 9.45 0.04496
        0.01139 11.45 0.04058
        0.01190 12.45 0.05986
        0.00359 13.45 0.04546
        0.05141 14.45 0.07268
        0.00472 15.45 0.07422
        0.00318 17.95 0.04007
        0.01270 19.95 0.05185
        0.01443 25.45 0.04049
      };
      \addplot[dHurtO, delta marks] table[x index=0, y index=1, x error index=2] {
        -0.04318 0.95 0.08167
        -0.08698 3.95 0.12098
        -0.01752 4.95 0.08096
        -0.02182 6.95 0.12174
        -0.00944 18.95 0.02145
        -0.02406 20.95 0.07519
        -0.02262 23.45 0.03349
        -0.01159 24.45 0.04079
        -0.00114 26.45 0.05665
      };
    \nextgroupplot[title={ReProbe}]
      \draw[black!8, line width=0.4pt] (axis cs:-0.40,0.95) -- (axis cs:0.40,0.95);
      \draw[black!8, line width=0.4pt] (axis cs:-0.40,1.95) -- (axis cs:0.40,1.95);
      \draw[black!8, line width=0.4pt] (axis cs:-0.40,2.95) -- (axis cs:0.40,2.95);
      \draw[black!8, line width=0.4pt] (axis cs:-0.40,3.95) -- (axis cs:0.40,3.95);
      \draw[black!8, line width=0.4pt] (axis cs:-0.40,4.95) -- (axis cs:0.40,4.95);
      \draw[black!8, line width=0.4pt] (axis cs:-0.40,5.95) -- (axis cs:0.40,5.95);
      \draw[black!8, line width=0.4pt] (axis cs:-0.40,6.95) -- (axis cs:0.40,6.95);
      \draw[black!8, line width=0.4pt] (axis cs:-0.40,9.45) -- (axis cs:0.40,9.45);
      \draw[black!8, line width=0.4pt] (axis cs:-0.40,10.45) -- (axis cs:0.40,10.45);
      \draw[black!8, line width=0.4pt] (axis cs:-0.40,11.45) -- (axis cs:0.40,11.45);
      \draw[black!8, line width=0.4pt] (axis cs:-0.40,12.45) -- (axis cs:0.40,12.45);
      \draw[black!8, line width=0.4pt] (axis cs:-0.40,13.45) -- (axis cs:0.40,13.45);
      \draw[black!8, line width=0.4pt] (axis cs:-0.40,14.45) -- (axis cs:0.40,14.45);
      \draw[black!8, line width=0.4pt] (axis cs:-0.40,15.45) -- (axis cs:0.40,15.45);
      \draw[black!8, line width=0.4pt] (axis cs:-0.40,17.95) -- (axis cs:0.40,17.95);
      \draw[black!8, line width=0.4pt] (axis cs:-0.40,18.95) -- (axis cs:0.40,18.95);
      \draw[black!8, line width=0.4pt] (axis cs:-0.40,19.95) -- (axis cs:0.40,19.95);
      \draw[black!8, line width=0.4pt] (axis cs:-0.40,20.95) -- (axis cs:0.40,20.95);
      \draw[black!8, line width=0.4pt] (axis cs:-0.40,23.45) -- (axis cs:0.40,23.45);
      \draw[black!8, line width=0.4pt] (axis cs:-0.40,24.45) -- (axis cs:0.40,24.45);
      \draw[black!8, line width=0.4pt] (axis cs:-0.40,25.45) -- (axis cs:0.40,25.45);
      \draw[black!8, line width=0.4pt] (axis cs:-0.40,26.45) -- (axis cs:0.40,26.45);
      \draw[black!45, line width=0.5pt] (axis cs:0,-0.75) -- (axis cs:0,27.45);
      \addplot[dHelpF, delta marks] table[x index=0, y index=1, x error index=2] {
        0.12831 1.95 0.10419
        0.11267 5.95 0.10518
        0.08301 12.45 0.08125
        0.22682 13.45 0.08304
        0.10134 17.95 0.03358
        0.09377 24.45 0.04864
      };
      \addplot[dHelpO, delta marks] table[x index=0, y index=1, x error index=2] {
        0.07888 0.95 0.08162
        0.00203 4.95 0.13974
        0.04338 9.45 0.07199
        0.04694 10.45 0.04897
        0.04764 11.45 0.06199
        0.04483 14.45 0.05545
        0.05965 15.45 0.06936
        0.04568 18.95 0.04572
        0.01636 19.95 0.04195
        0.03005 23.45 0.03559
        0.05346 25.45 0.06563
        0.03660 26.45 0.05200
      };
      \addplot[dHurtO, delta marks] table[x index=0, y index=1, x error index=2] {
        -0.01118 2.95 0.09815
        -0.14009 3.95 0.21690
        -0.02222 6.95 0.07027
        -0.00696 20.95 0.08754
      };
    \end{groupplot}
  \end{tikzpicture}
  \caption{$\Delta A_{c}$ of Section~\ref{sec:design} for every generator--subset combination, with its solution count. A bar is the standard deviation over the 10 seeds of the within-seed difference. A marker is blue when the change is positive, red when it is negative and gray when it is zero, and filled when the change is larger in magnitude than its bar.}
  \label{fig:delta}
\end{figure}

Table~\ref{tab:main} gives, for each method without and with curvature, the mean of $A$ over the generator--subset combinations of each subset with its standard error over the generator--subset combinations, and Tables~\ref{tab:full_gsm8k} to~\ref{tab:full_omnimath} give $A$ for every generator--subset combination and configuration with its standard deviation over the 10 seeds, and Figure~\ref{fig:delta} gives the paired difference of every generator--subset combination with its standard deviation over the seeds. Table~\ref{tab:attention_controls} gives the corresponding subset means for the two attention controls. Every value is from our own runs. The values use ProcessBench's definition, which that benchmark calls the F1 score, but are not comparable to the values it publishes, which pool a subset's generators.

\begin{table}[htbp]
    \caption{Mean $A$ over the generator--subset combinations of each subset, weighted by each generator--subset combination's solution count, with the standard error of that mean over the generator--subset combinations in parentheses, the weighted sample standard deviation of the generator--subset combination values divided by the square root of the Kish effective number of generator--subset combinations. The four subsets contain 7, 7, 4, and 4 generator--subset combinations. Each method appears twice, without $\mathbf{q}$ and then with $\mathbf{q}$ appended to its input. \emph{All} is the same weighted mean over all 22 generator--subset combinations, the value Table~\ref{tab:pairs} reports. Tables~\ref{tab:full_gsm8k} to~\ref{tab:full_omnimath} give each generator--subset combination separately.}
    \label{tab:main}
    \centering
    \small
    \setlength{\tabcolsep}{5pt}
    \begin{tabular}{@{}lcccc@{\hspace{1.4em}}c@{}}
        \toprule
        Method & GSM8K & MATH & OlympiadBench & Omni-MATH & All \\
        \midrule
        Spectral, logistic & 0.279\,{\scriptsize(0.045)} & 0.302\,{\scriptsize(0.024)} & 0.328\,{\scriptsize(0.021)} & 0.377\,{\scriptsize(0.014)} & 0.326\,{\scriptsize(0.016)} \\
        \quad with curvature & 0.321\,{\scriptsize(0.045)} & 0.337\,{\scriptsize(0.024)} & 0.338\,{\scriptsize(0.015)} & 0.383\,{\scriptsize(0.014)} & 0.347\,{\scriptsize(0.014)} \\
        Hidden states, linear & 0.096\,{\scriptsize(0.044)} & 0.143\,{\scriptsize(0.021)} & 0.198\,{\scriptsize(0.033)} & 0.261\,{\scriptsize(0.035)} & 0.183\,{\scriptsize(0.021)} \\
        \quad with curvature & 0.111\,{\scriptsize(0.035)} & 0.158\,{\scriptsize(0.024)} & 0.200\,{\scriptsize(0.030)} & 0.262\,{\scriptsize(0.042)} & 0.191\,{\scriptsize(0.020)} \\
        MLP probe & 0.187\,{\scriptsize(0.041)} & 0.249\,{\scriptsize(0.029)} & 0.340\,{\scriptsize(0.040)} & 0.376\,{\scriptsize(0.041)} & 0.299\,{\scriptsize(0.024)} \\
        \quad with curvature & 0.181\,{\scriptsize(0.035)} & 0.274\,{\scriptsize(0.031)} & 0.336\,{\scriptsize(0.039)} & 0.370\,{\scriptsize(0.037)} & 0.304\,{\scriptsize(0.023)} \\
        ReProbe & 0.210\,{\scriptsize(0.054)} & 0.254\,{\scriptsize(0.028)} & 0.317\,{\scriptsize(0.021)} & 0.360\,{\scriptsize(0.026)} & 0.293\,{\scriptsize(0.021)} \\
        \quad with curvature & 0.238\,{\scriptsize(0.057)} & 0.327\,{\scriptsize(0.014)} & 0.362\,{\scriptsize(0.030)} & 0.413\,{\scriptsize(0.022)} & 0.347\,{\scriptsize(0.020)} \\
        \bottomrule
    \end{tabular}
\end{table}

\begin{table}[htbp]
    \caption{Every method on the GSM8K subset, by generator, in its configuration without curvature and with the curvature vector $\mathbf{q}$ appended (+curv). Values are $A$, the mean over the 10 seeds, with the sample standard deviation over the seeds beneath. $n$ is the generator--subset combination's solution count. \texttt{SpL} spectral logistic, \texttt{Hid} hidden states linear, \texttt{MLP} the MLP probe, \texttt{RePr} ReProbe.}
    \label{tab:full_gsm8k}
    \centering
    \scriptsize
    \setlength{\tabcolsep}{2.6pt}
    \begin{tabular}{lrcccccccc}
        \toprule
        Generator & $n$ & SpL & SpL+curv & Hid & Hid+curv & MLP & MLP+curv & RePr & RePr+curv \\
        \midrule
        \texttt{Llama-3.1-8B} & 61 & 0.391 & 0.437 & 0.045 & 0.107 & 0.271 & 0.228 & 0.219 & 0.297 \\
         & & {\tiny(0.062)} & {\tiny(0.052)} & {\tiny(0.049)} & {\tiny(0.058)} & {\tiny(0.039)} & {\tiny(0.096)} & {\tiny(0.079)} & {\tiny(0.061)} \\
        \texttt{Meta-Llama-3-8B} & 24 & 0.098 & 0.273 & 0.000 & 0.016 & 0.090 & 0.142 & 0.000 & 0.113 \\
         & & {\tiny(0.142)} & {\tiny(0.105)} & {\tiny(0.000)} & {\tiny(0.052)} & {\tiny(0.123)} & {\tiny(0.085)} & {\tiny(0.000)} & {\tiny(0.105)} \\
        \texttt{Qwen2-1.5B} & 41 & 0.306 & 0.312 & 0.295 & 0.251 & 0.302 & 0.215 & 0.372 & 0.232 \\
         & & {\tiny(0.137)} & {\tiny(0.145)} & {\tiny(0.065)} & {\tiny(0.065)} & {\tiny(0.057)} & {\tiny(0.141)} & {\tiny(0.075)} & {\tiny(0.163)} \\
        \texttt{Qwen2-7B} & 52 & 0.330 & 0.405 & 0.079 & 0.114 & 0.134 & 0.182 & 0.293 & 0.421 \\
         & & {\tiny(0.044)} & {\tiny(0.033)} & {\tiny(0.056)} & {\tiny(0.042)} & {\tiny(0.055)} & {\tiny(0.081)} & {\tiny(0.101)} & {\tiny(0.096)} \\
        \texttt{Qwen2.5-1.5B} & 42 & 0.254 & 0.234 & 0.164 & 0.157 & 0.233 & 0.277 & 0.262 & 0.251 \\
         & & {\tiny(0.060)} & {\tiny(0.050)} & {\tiny(0.067)} & {\tiny(0.059)} & {\tiny(0.094)} & {\tiny(0.059)} & {\tiny(0.057)} & {\tiny(0.064)} \\
        \texttt{Qwen2.5-7B} & 27 & 0.264 & 0.284 & 0.000 & 0.000 & 0.060 & 0.042 & 0.044 & 0.046 \\
         & & {\tiny(0.124)} & {\tiny(0.177)} & {\tiny(0.000)} & {\tiny(0.000)} & {\tiny(0.077)} & {\tiny(0.094)} & {\tiny(0.098)} & {\tiny(0.074)} \\
        \texttt{Qwen2.5-Math-7B} & 22 & 0.064 & 0.085 & 0.000 & 0.000 & 0.043 & 0.021 & 0.022 & 0.000 \\
         & & {\tiny(0.103)} & {\tiny(0.109)} & {\tiny(0.000)} & {\tiny(0.000)} & {\tiny(0.091)} & {\tiny(0.067)} & {\tiny(0.070)} & {\tiny(0.000)} \\
        \bottomrule
    \end{tabular}
\end{table}

\begin{table}[htbp]
    \caption{As Table~\ref{tab:full_gsm8k}, for the MATH subset.}
    \label{tab:full_math}
    \centering
    \scriptsize
    \setlength{\tabcolsep}{2.6pt}
    \begin{tabular}{lrcccccccc}
        \toprule
        Generator & $n$ & SpL & SpL+curv & Hid & Hid+curv & MLP & MLP+curv & RePr & RePr+curv \\
        \midrule
        \texttt{Llama-3.1-8B} & 139 & 0.306 & 0.334 & 0.203 & 0.214 & 0.350 & 0.368 & 0.314 & 0.357 \\
         & & {\tiny(0.036)} & {\tiny(0.051)} & {\tiny(0.053)} & {\tiny(0.033)} & {\tiny(0.043)} & {\tiny(0.049)} & {\tiny(0.049)} & {\tiny(0.046)} \\
        \texttt{Meta-Llama-3-8B} & 70 & 0.391 & 0.437 & 0.198 & 0.220 & 0.294 & 0.345 & 0.287 & 0.332 \\
         & & {\tiny(0.049)} & {\tiny(0.020)} & {\tiny(0.021)} & {\tiny(0.038)} & {\tiny(0.075)} & {\tiny(0.062)} & {\tiny(0.048)} & {\tiny(0.039)} \\
        \texttt{Qwen2-1.5B} & 47 & 0.347 & 0.372 & 0.186 & 0.259 & 0.303 & 0.307 & 0.291 & 0.350 \\
         & & {\tiny(0.047)} & {\tiny(0.054)} & {\tiny(0.088)} & {\tiny(0.075)} & {\tiny(0.067)} & {\tiny(0.061)} & {\tiny(0.048)} & {\tiny(0.059)} \\
        \texttt{Qwen2-7B} & 131 & 0.358 & 0.389 & 0.101 & 0.113 & 0.224 & 0.280 & 0.284 & 0.331 \\
         & & {\tiny(0.048)} & {\tiny(0.035)} & {\tiny(0.042)} & {\tiny(0.046)} & {\tiny(0.036)} & {\tiny(0.025)} & {\tiny(0.032)} & {\tiny(0.044)} \\
        \texttt{Qwen2.5-1.5B} & 74 & 0.229 & 0.278 & 0.120 & 0.120 & 0.202 & 0.206 & 0.100 & 0.327 \\
         & & {\tiny(0.085)} & {\tiny(0.069)} & {\tiny(0.045)} & {\tiny(0.046)} & {\tiny(0.061)} & {\tiny(0.062)} & {\tiny(0.059)} & {\tiny(0.047)} \\
        \texttt{Qwen2.5-7B} & 97 & 0.238 & 0.277 & 0.141 & 0.145 & 0.210 & 0.221 & 0.209 & 0.256 \\
         & & {\tiny(0.044)} & {\tiny(0.053)} & {\tiny(0.045)} & {\tiny(0.038)} & {\tiny(0.042)} & {\tiny(0.018)} & {\tiny(0.045)} & {\tiny(0.043)} \\
        \texttt{Qwen2.5-Math-7B} & 96 & 0.256 & 0.287 & 0.069 & 0.088 & 0.155 & 0.167 & 0.250 & 0.333 \\
         & & {\tiny(0.074)} & {\tiny(0.093)} & {\tiny(0.038)} & {\tiny(0.058)} & {\tiny(0.029)} & {\tiny(0.055)} & {\tiny(0.054)} & {\tiny(0.050)} \\
        \bottomrule
    \end{tabular}
\end{table}

\begin{table}[htbp]
    \caption{As Table~\ref{tab:full_gsm8k}, for the OlympiadBench subset.}
    \label{tab:full_olympiadbench}
    \centering
    \scriptsize
    \setlength{\tabcolsep}{2.6pt}
    \begin{tabular}{lrcccccccc}
        \toprule
        Generator & $n$ & SpL & SpL+curv & Hid & Hid+curv & MLP & MLP+curv & RePr & RePr+curv \\
        \midrule
        \texttt{Llama-3.1-8B} & 164 & 0.346 & 0.348 & 0.269 & 0.258 & 0.391 & 0.394 & 0.293 & 0.394 \\
         & & {\tiny(0.023)} & {\tiny(0.032)} & {\tiny(0.040)} & {\tiny(0.036)} & {\tiny(0.031)} & {\tiny(0.043)} & {\tiny(0.033)} & {\tiny(0.026)} \\
        \texttt{Qwen2-7B} & 108 & 0.256 & 0.294 & 0.141 & 0.139 & 0.292 & 0.268 & 0.314 & 0.307 \\
         & & {\tiny(0.051)} & {\tiny(0.033)} & {\tiny(0.064)} & {\tiny(0.059)} & {\tiny(0.035)} & {\tiny(0.061)} & {\tiny(0.031)} & {\tiny(0.072)} \\
        \texttt{Qwen2.5-7B} & 123 & 0.339 & 0.332 & 0.136 & 0.143 & 0.237 & 0.250 & 0.283 & 0.300 \\
         & & {\tiny(0.033)} & {\tiny(0.037)} & {\tiny(0.034)} & {\tiny(0.038)} & {\tiny(0.037)} & {\tiny(0.038)} & {\tiny(0.032)} & {\tiny(0.028)} \\
        \texttt{Qwen2.5-Math-7B} & 147 & 0.352 & 0.366 & 0.211 & 0.228 & 0.403 & 0.393 & 0.373 & 0.419 \\
         & & {\tiny(0.037)} & {\tiny(0.033)} & {\tiny(0.039)} & {\tiny(0.036)} & {\tiny(0.026)} & {\tiny(0.026)} & {\tiny(0.046)} & {\tiny(0.035)} \\
        \bottomrule
    \end{tabular}
\end{table}

\begin{table}[htbp]
    \caption{As Table~\ref{tab:full_gsm8k}, for the Omni-MATH subset.}
    \label{tab:full_omnimath}
    \centering
    \scriptsize
    \setlength{\tabcolsep}{2.6pt}
    \begin{tabular}{lrcccccccc}
        \toprule
        Generator & $n$ & SpL & SpL+curv & Hid & Hid+curv & MLP & MLP+curv & RePr & RePr+curv \\
        \midrule
        \texttt{Llama-3.1-8B} & 162 & 0.406 & 0.402 & 0.342 & 0.356 & 0.473 & 0.450 & 0.426 & 0.456 \\
         & & {\tiny(0.029)} & {\tiny(0.035)} & {\tiny(0.041)} & {\tiny(0.032)} & {\tiny(0.035)} & {\tiny(0.017)} & {\tiny(0.033)} & {\tiny(0.029)} \\
        \texttt{Qwen2-7B} & 131 & 0.376 & 0.403 & 0.270 & 0.274 & 0.361 & 0.375 & 0.328 & 0.381 \\
         & & {\tiny(0.041)} & {\tiny(0.035)} & {\tiny(0.044)} & {\tiny(0.048)} & {\tiny(0.028)} & {\tiny(0.036)} & {\tiny(0.026)} & {\tiny(0.062)} \\
        \texttt{Qwen2.5-7B} & 104 & 0.338 & 0.344 & 0.191 & 0.201 & 0.279 & 0.278 & 0.326 & 0.362 \\
         & & {\tiny(0.050)} & {\tiny(0.045)} & {\tiny(0.043)} & {\tiny(0.054)} & {\tiny(0.055)} & {\tiny(0.072)} & {\tiny(0.032)} & {\tiny(0.050)} \\
        \texttt{Qwen2.5-Math-7B} & 132 & 0.374 & 0.368 & 0.210 & 0.183 & 0.349 & 0.337 & 0.336 & 0.430 \\
         & & {\tiny(0.041)} & {\tiny(0.030)} & {\tiny(0.034)} & {\tiny(0.035)} & {\tiny(0.025)} & {\tiny(0.039)} & {\tiny(0.054)} & {\tiny(0.044)} \\
        \bottomrule
    \end{tabular}
\end{table}

\begin{table}[htbp]
    \caption{Mean $A$ for the two attention controls, averaged over the generator--subset combinations of each subset with solution-count weights. Each combination is first averaged over the ten seeds. The All column uses solution-count weights over all 22 combinations. Table~\ref{tab:main} gives the corresponding values without added features and with curvature.}
    \label{tab:attention_controls}
    \centering
    \small
    \setlength{\tabcolsep}{4pt}
    \begin{tabular}{@{}llccccc@{}}
        \toprule
        Method & Control & GSM8K & MATH & OlympiadBench & Omni-MATH & All \\
        \midrule
        Spectral, logistic & Neg-log & 0.3363 & 0.2840 & 0.3141 & 0.3490 & 0.3165 \\
        Spectral, logistic & Stats & 0.3299 & 0.2882 & 0.3371 & 0.3738 & 0.3298 \\
        Hidden states, linear & Neg-log & 0.1018 & 0.1469 & 0.2032 & 0.2640 & 0.1872 \\
        Hidden states, linear & Stats & 0.0977 & 0.1514 & 0.1966 & 0.2578 & 0.1847 \\
        MLP probe & Neg-log & 0.1752 & 0.2655 & 0.3384 & 0.3792 & 0.3033 \\
        MLP probe & Stats & 0.1806 & 0.2507 & 0.3505 & 0.3776 & 0.3021 \\
        ReProbe & Neg-log & 0.1922 & 0.2814 & 0.3176 & 0.3853 & 0.3068 \\
        ReProbe & Stats & 0.1999 & 0.2856 & 0.3275 & 0.3795 & 0.3103 \\
        \bottomrule
    \end{tabular}
\end{table}

\end{document}